# Land Use Classification Using Multi-neighborhood LBPs

**Harjot Singh Parmar,**
Systems Design Engineering Dept.
*University Of Waterloo*
Ontario, Canada
hsparmar@uwaterloo.ca

*Abstract*— **In this paper we propose the use of multiple local binary patterns(LBPs) to effectively classify land use images. We use the UC Merced 21 class land use image dataset. Task is challenging for classification as the dataset contains intra class variability and inter class similarities. Our proposed method of using multi-neighborhood LBPs combined with nearest neighbor classifier is able to achieve an accuracy of 77.76%. Further class wise analysis is conducted and suitable suggestion are made for further improvements to classification accuracy.**

*Keywords— Computer Vision, Pattern Recognition, Remote Sensing, Land Use Classification.*

I. INTRODUCTION

The world is changing rapidly, new technology and infrastructure is resulting in faster growth. To meet the demands of the growing populations, cities are expanding and land use pattern are changing to accommodate the needs. Due to these developments the need to constantly monitor and assess these changes is increasing.

Land use classification provides information on land cover, and the types of human activity involved in land use. It also facilitates the assessment of environmental impacts on, and potential or alternative uses of land.

By using computational techniques to observe temporal changes in land-use patterns, land use classification helps prevent problems associated with haphazard and uncontrolled land development, deteriorating environmental quality, loss of prime agricultural lands, destruction of important wetlands, loss of fish and wildlife habitat. [1]

Automated land-use classification has become highly desirable due to availability of high-resolution remote sensing images. Since these images poses rich textural information, they can be used to differentiate between the land-use patterns, approaches which use texture information are widely used in this domain.

In order to generate effective insights from the images, it is paramount to select effective feature descriptors, which could be later trained using suitable classifiers, to accurately tag unlabeled images.

In this paper we propose the use of Local Binary Patterns(LBPs) and Multi-neighborhood LBP descriptors combined with K-nearest neighbor classifier to identify images in the UC Merced land use dataset. Local binary patterns were first introduced by Ojala et al in 1994.[2] The first proposed method used a simple neighborhood thresholding to generate features. A more robust grey-scale and rotation invariant version was introduced in 2002. [3] Ability to capture image texture, ease of computation and simplicity for interpretation makes use of LBP features suitable for land-use classification problem. LBP have previously been effectively used for classification of histopathological images [4]. In case of histopathological images, classes contain intra and inter-class similarities, which makes classification challenging. LBP have also been used for age detection using facial images. [11]

This paper is structured in the following fashion: In section (II) we discuss some of related work on the same dataset. Section(III) and (IV) provides background and understanding about local binary patterns and nearest neighborhood classifier respectively. Section(V) we describe the dataset. Followed by description of Method/Approach in section (VI). Testing/ Validation approach is explained in section (VII). We comment on the experimental findings and results in section (VIII) followed by suggestions/recommendations in section (IX) and finally draw conclusion in section (X).

II. RELATED WORK

In the recent years, there has been tremendous growth in the research around remote sensing scene recognition. UC Merced dataset has been the most widely used dataset for evaluating and comparing the effectiveness of different techniques. Earliest research literature effectively used hand crafted features in combination with a classifier. Lately artificial neural networks have gained popularity having shown impressive results in object classification[7]. Hence, ANN have also been used in remote sensing and land use classification. In this section we will briefly discuss some of the methods used for land use classification.

Yang and N. Newsam[5] introduced UC Merced dataset in 2010, and in this paper, the first benchmark was set using Bag Of Visual Words(BOVW) method. BOVW is a commonly used technique. In BOVW, first K means clustering is used to generate an off-line dictionary of visual words. This dictionary is then used to quantize the descriptors extracted from the images by associating the descriptors with the nearest cluster label. Finally, the histogram of these labels is fed to a classifier to generate the model. Yan and N.Newsam used SIFT and BOVW in combination with the SVM to achieve an accuracy of 76.81%. Performance of BOVW is limited due to fact that the basic implementation of BOVW doesn't take into consideration the spatial distribution of the visual words. A more robust approach is spatial pyramid match kernels [6] which incorporates image partitioning at multiple resolutions and then computes histogram of features in each of the partition to produce a descriptor. Which can be used with a classifier to generate model.

Even though the images in the dataset are colored images. In our method, we are using gray-scale images to extract features. RGB bands have also been used in M-centrist, [9] a multi-channel approach derived from Centrist approach, [10] which uses LBP and PCA to reduce the feature vector dimensionality. M-centerist method achieved an accuracy of 89.90%. Yang and Newsam [5] have also successfully demonstrated the use of color histogram descriptors (Color-HLS) for the task, achieving an accuracy of 81.19%.

Neural networks take inspiration from the human brain, and try to replicate the complex decision making by using an architecture comprising of multiple layers of artificial neurons. Use and interest in convolutional Neural Networks has grown rapidly in the domain of image classification and retrieval due to availability of better hardware and introduction of new libraries such as TensorFlow, Keras etc. There are couple ways you can use neural networks to perform classification. The most fundamental being training a completely new network based on the dataset, but this can computationally expensive and time consuming. The other method is to use a pretrained neural nets: either tune it to fit your dataset or use the output from the penultimate layer to generate features for your images, which can be used with an offline classifier to generate the model. In [12] they used both a neural networks trained from scratch and a fine-tuned neural network model to achieve an accuracy of 92.86% and 97.10% respectively. In [8] a pretrained neural network was used to generate shallow feature vectors for SVM classification, achieving an accuracy of 93.42%.

### III. LOCAL BINARY PATTERNS

Local binary patterns were first introduced in 1994 by Ojala et al. They are used as image descriptors, to uniquely define images based on the greyscale value thresholding.

The easiest way to extract a local binary pattern(LBP) feature vector for a given image is as follows:

- For each pixel in an image consider a neighborhood, eg 3x3 window neighborhood.
- For a given 3x3 window, the central pixel will have 8 neighboring pixels.
- Now, traversing in a clockwise or counter-clockwise fashion, we threshold the neighboring pixels against the central pixel and generate a binarized descriptor for the particular central pixel.
- Once we have generated the binarized descriptor for each pixel, we compute the histogram of these descriptors over the whole image. The histogram thus generated represents the feature vector for that particular image.

Now, even if we consider 8 neighboring pixels, it evaluates to 2^8 = 256 possible values for each pixel in the image. With increase in image size or the number of neighborhood pixels considered, computational cost increases exponentially.

So in 2002, Ojala et al [3] introduced Uniform LBP, which performed lossy encoding, helping eliminate some irrelevant binary values and retain the useful ones. Uniform LBP are identified as the binary descriptors which contains at most two 0-1 or 1-0 transitions.

These Uniform LBP are interpreted as lines, corners or ramp gradient. The pattern of LBP which contain more than 2 transitions are considered as "Non-uniform" LBP. Histogram is generated based on these unique uniform patterns and considering all other non-uniform LBPs as one entity. This allows for faster computation of the feature vectors. Uniform LBPs allows for finer quantization of angular space, they are also grayscale and rotation invariant. *Figure 1.* shows the different possible patterns in LBPs, black and white dots represent the thresholded pixels.

In order to obtain good results, parameters such as the distance from the central pixel(radius) and number of neighborhood pixels are considered for computing of LBP. LBPs can also be calculated to be rotation variant or grayscale variant.

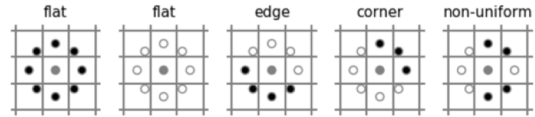

*Figure 1 Possible LBP patterns in 3X3 neighborhood and their interpretation. White dots represent 1, black dot represent 0 in the pattern.*

### IV. NEAREST NEIGHBOR CLASSIFIER

Nearest Neighbor Classifier is a non-parameteric classifier or non-generalizing classifier. What that means is that it doesn't construct a model or rules to generate predictions. In classification, class is assigned based on the weighted votes of the nearest neighbors. Nearest point selection can be based on the preselected count of points (K) or based on the radius distance, within which points are considered. Optimal value of K is based on heuristic and is data dependent. Ideally larger K value, suppresses the noise, as more number of points are considered, which deals with the chance of witnessing an outlier, nearest to the point in consideration. Radius based point selection is used when the data is not uniformly sampled.

As class assignment is dependent on the votes of the neighborhood points. Weights can be assigned to these votes. Generally, uniform weights are used, which assigns equal weights to the points considered in the neighborhood. Weights can be assigned using the distance of the points as the parameter, in cases where, it might be required to assign higher weight to the points nearer to the query point.

### V. DATASET

The UC Merced land-use dataset is [5] used for this project. The dataset was extracted from United States Geological Survey national maps. Dataset consists of 21 land-use classes, extracted by manually selecting 100 images for each class. Each image is 256x256 pixels in size. The classes are labeled under the categories mentioned below:

- agricultural
- airplane
- baseball diamond
- beach
- buildings
- chaparral
- dense residential
- forest
- freeway
- golf course

- harbor
- intersection
- medium residential
- mobile homepark
- overpass
- parking lot
- river
- runway
- sparse residential
- storage tanks
- tennis court

As we will be working with greyscale images. *Figure 2*. shows the grayscale samples of each class with the associated label. The classes have good inter and intra class variability, which poses as a challenge for classifiers. The dataset has been used to evaluate classifier performance in a number of publications [8][9][12].

## VI. METHOD / APPROACH

In order to classify images, we first converted all the 2100 images to grayscale. We decided to use a multiple combination of radius (r) and number of neighborhood points (p) to generate LBP feature vectors. The combinations (r,p) used are (8,1), (16,2), (24,3), (32,4), (32,5). This combination allows us to generate feature vectors at a micro level ie.1-2 pixel distance and macro level feature at a distance of 3-5 pixels. *Figure 3.* shows the LBP representation in these neighborhood for a dense residential class.

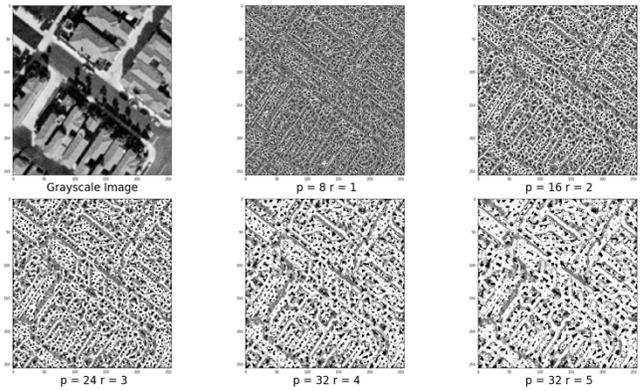

*Figure 3. LBP representation of dense residential class for different combination of LBP parameters.*

For each of the radius and number of neighborhood points pair, we generated a master dataset containing the class labels and their respective LBP feature vectors. These feature vectors were then used to train and test classifier and generate predictions.

Mutli-class classifiers such as Random forest, Decision tree, naïve bayes and K-nearest neighbors were initially tested for classification. Based on the performance, K-nearest neighbors was chosen for further analysis.

We used K = 1, 3 and 5 as the number of the neighborhood points considered for votes. Votes were weighted based on two distance measures are defined below.

- Minkowski distance:

Given two points $X_i = (x_{i1}, x_{i2}, x_{i3...})$ and $X_j = (x_{j1}, x_{j2}, x_{j3...})$, we calculate minkowski distance according to the eqution given below. For our experiment we are using order (p) = 2, which is in effect euclidean distance.

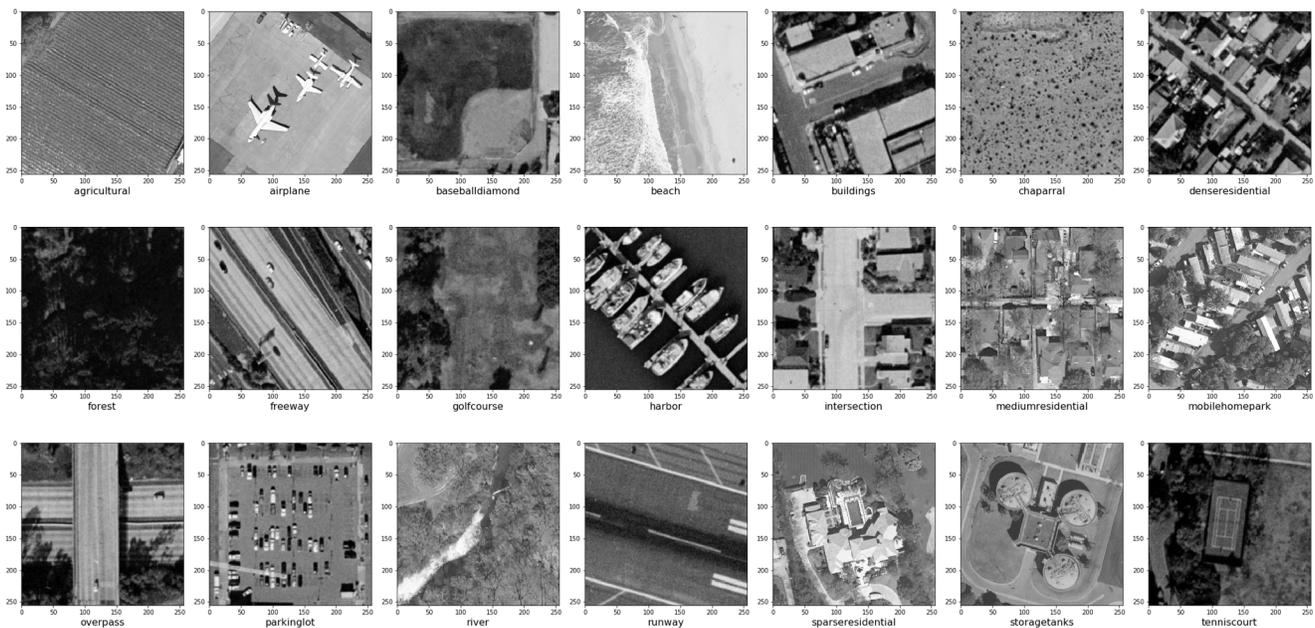

*Figure 2 Sample Grayscale Images of each class in UC Merced Dataset*

$$D(\mathbf{x}_i, \mathbf{x}_j) = \left( \sum_{l=1}^{d} |x_{il} - x_{jl}|^p \right)^{1/p}$$

-- Minkowski Distance

- Chi-square Distance:

Given two points $P = (p_1, p_2, p_3...)$ and $Q = (q_1, q_2, q_3...)$ Chi-square distance is calculated as according to the formula given below. Chi-square is one of the popular distance measures used with LBPs, and have been observed to give good results.

$$\chi^2_{A,B} = \frac{1}{2} \sum_{i=1}^{n} \left( \frac{[p(i) - q(i)]^2}{p(i) + q(i)} \right)$$

-- Chi-Square Distance

Post the preliminary classification tests with singular LBPs we were able to establish the performance at a neighborhood level. We then used combination of these singular LBPs to generate multi-neighborhood LBPs. For example, by concatenating the feature vector from (8,1) and (24, 3) we were able to generate a feature vector which combined both the micro and macro level features, which would be more representative of the overall image. Classifier were again trained and tested in the same fashion as the done previously with the singular neighborhood LBPs.

## VII. TESTING / VALIDATION

Testing was carried out using two method listed below:

a) K fold cross-validation

In K-fold cross validation we split the data into K-different subsets and use the K-1 subsets to train and last subset is kept for testing/validation. The process is repeated for the K folds or times, using each subset once for testing. Accuracy is averaged out over the K-folds, to determine the performance of the classifier.

b) Leave One Out (LOO)

In Leave One Out(LOO) methodology, number of subsets is equal to the number of observations in the dataset. Each observation is left out as the test and rest of the dataset is used for training purposes, process is repeated all the observations. Again, we average out the prediction performance on all the folds to determine the accuracy.

For initial testing we used the K-fold cross validation to measure the performance, as LOO, would have been computationally time consuming, for preliminary experiments. We used 10 fold cross-validation for initial accuracy measures, which gave a directional idea. This was carried out for both the distance measures. After establishing that Chi-square distance gave better results. We ran LOO accuracy measurement for the singular LBP based on Chi-square distance.

After running cross validation on singular LBPs, Accuracy of multi-neighborhood LBPs was evaluated in the similar fashion, we only used LOO methodology for this phase of experiments.

## VIII. RESULTS AND DISCUSSIONS

The experiments were carried out on a 8GB RAM, 2.6GHz Intel i5 core system. Algorithms were implemented using Python programming language.

Initial experimental results using 10 fold cross validation are presented in the *Table 1*. Accuracies using the chi-square distance were observed to be on average in the higher 60% region, The 5-nearest neighbor classifier based on LBP(32,4) feature vector achieved the highest accuracy of 68.84%. Whereas using Euclidean distance (*Table 2.*) we were able to achieve accuracies in the mid 50s and Low 60s. Here, 3-nearest neighbor classifier based on the LBP(8,1) feature vector achieved the highest accuracy of 61.25%. Due to considerable difference in the average accuracies of two distance measures, we decided to continue experiments using chi-square distance only.

| LBP(Points, Radius) | Chi-square (10 Fold CV) | | |
|---|---|---|---|
| | K = 1 | K = 3 | K = 5 |
| (8,1) | 64.40 | 65.17 | 65.06 |
| (16,2) | 66.29 | 66.23 | 65.73 |
| (24,3) | 66.17 | 67.55 | 66.21 |
| (32,4) | 65.67 | 68.44 | **68.64** |
| (32,5) | 63.52 | 63.67 | 64.90 |

Table 1. 10 fold cross-validation accuracy results (%) using chi-square distance for different LBP parameter combination.

| LBP (Points, Radius) | Euclidean (10 Fold CV) | | |
|---|---|---|---|
| | K = 1 | K = 3 | K = 5 |
| (8,1) | 60.66 | **61.25** | 59.23 |
| (16,2) | 59.23 | 59.28 | 60.42 |
| (24,3) | 58.45 | 57.44 | 57.12 |
| (32,4) | 57.24 | 57.34 | 56.23 |
| (32,5) | 54.71 | 54.67 | 54.67 |

Table 2. 10 fold cross-validation accuracy results (%) using Euclidean distance for different LBP parameter combination.

To substantiate our results, we repeated the experiments using Leave One Out methodology considering on Chi-square as the distance measure. Again we observed accuracies in range of 63-68%. With 5-nearest neighbor classifier using the (32,4) feature vector achieving the highest accuracy of 67.90%. *Table.3.* presents the results of the given experiment.

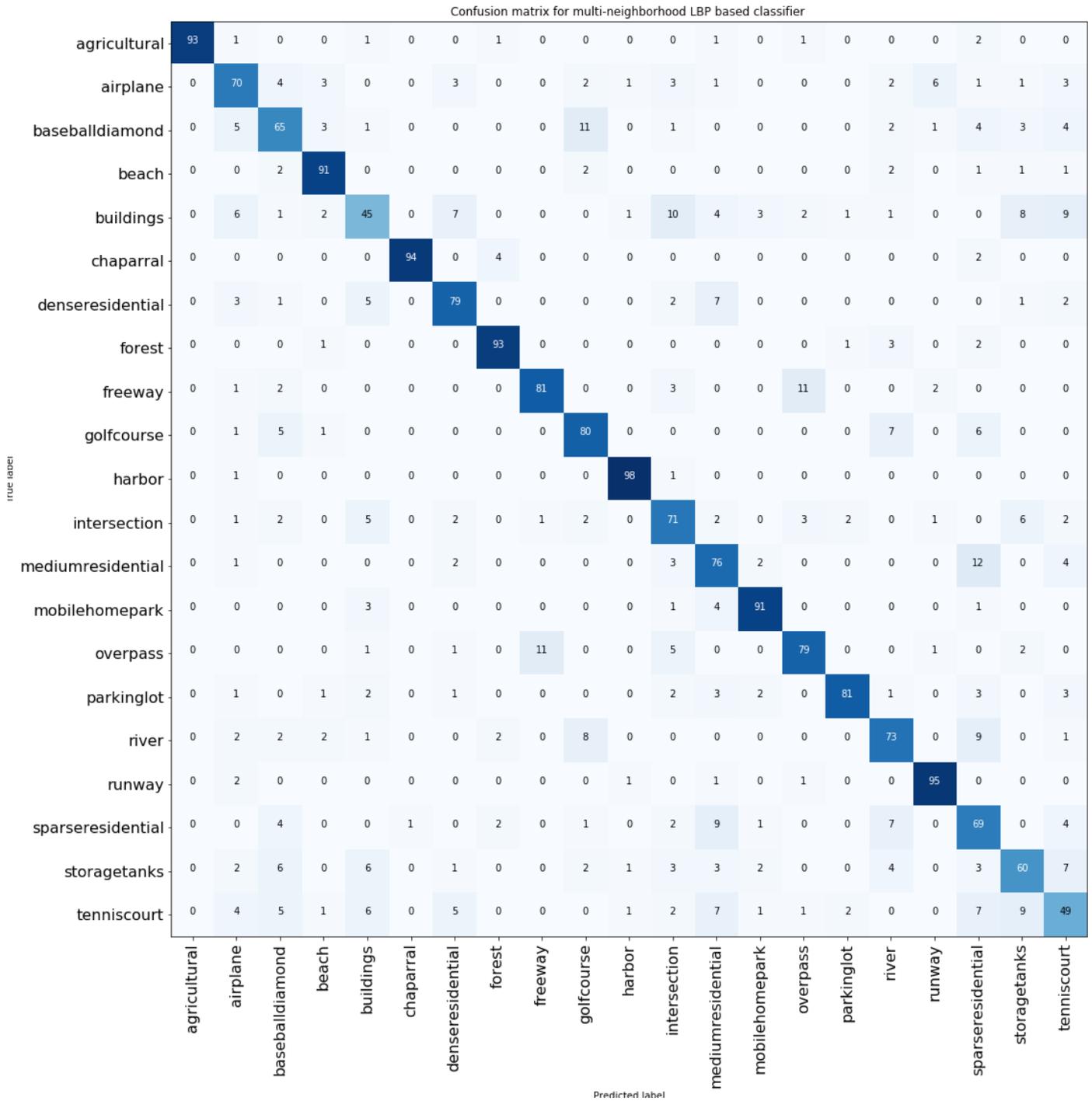

*Figure 4 Confusion matrix for multi-neighborhood classifier using (8,1)+(24,3)+(32,5) LBP.*

| LBP(Points, Radius) | Chi-square (LOO) | | |
|---|---|---|---|
| | K = 1 | K = 3 | K = 5 |
| (8,1) | 64.33 | 64.48 | 65.19 |
| (16,2) | 67.71 | 67.24 | 67.86 |
| (24,3) | 66.19 | 66.52 | 66.05 |
| (32,4) | 65.48 | 67.62 | **67.90** |
| (32,5) | 63.24 | 63.29 | 62.14 |

Table 3. Leave one out validation accuracies (%) using chi-square distance for different LBP parameter combination.

Next we ran experiments using a combination of multi-neighborhood LBPs with the same variation classifier. *Table 4.* presents the classification accuracies of the combinations using the Chi-square distance measure. We observed significant improvement in performance compared to singular LBPs, achieving accuracy improvement from mid 60% to mid and higher 70% region. This improvement indicates that by concatenating multi-neighborhood LBP descriptors we are able capture the properties of the image at a micro and macro level, augmented information thereby helped improve classifiers performance.

The model generated by concatenating the (8,1), (24,3) and (35,5) produced the best results. It performed slightly better than the benchmark in the initial BOVW method, [5] with an overall accuracy of 77.76% as against 76.81% of BOVW. Further a high kappa value of 0.76, corroborated the classifiers performance.

| Multi-neighborhood LBP(Points, Radius) | Chi-square (LOO) | | |
|---|---|---|---|
| | K = 1 | K = 3 | K = 5 |
| (8,1) + (16,2) | 72.29 | 72.10 | 71.62 |
| (8,1) + (16,2)+ (24,3) + (32,4) | 76.33 | 75.62 | 75.12 |
| (8,1) + (16,2)+ (24,3) + (32,4) + (32,5) | 77.29 | 76.19 | 75.48 |
| (8,1) + (24,3) + (32,5) | **77.76** | 76.90 | 75.67 |

Table 4. Leave one out validation accuracies (%) using chi-square distance for different LBP parameter combination.

To analyze the best performing model ((8,1) + (24,3) + (32,5) with 1-nearest neighbor classifier), we generated confusion matrix report to understand the classification performance at a per class level. *Figure 4.* Presents the confusion matrix report. Further we generated the class-wise classification report, recording the precision and recall values for each class. Results are presented in *Table 5*.

Analyzing both the reports we observed that the model was highly successful in identifying the agriculture, beach, chaparral, forest, habor, mobile parks and runway classes with high precision and recall values in the range of 0.85-1.00. Model had a hard time differentiating and perfomed poorly in case of tenniscourt, buildings, baseball diamonds and storage tanks, misclassifying close to 40% of the observations. Overpass and freeway classes were also highly confusing, each presenting with 11 instance where one was classified as the other. Similar confusion trend was observed in sparse, medium and dense residential.

| Class | Precision | Recall |
|---|---|---|
| agricultural | 1.00 | 0.93 |
| airplane | 0.69 | 0.70 |
| baseballdiamond | 0.66 | 0.65 |
| beach | 0.87 | 0.91 |
| buildings | 0.59 | 0.45 |
| chaparral | 0.99 | 0.94 |
| Dense residential | 0.78 | 0.79 |
| forest | 0.91 | 0.93 |
| freeway | 0.87 | 0.81 |
| golfcourse | 0.74 | 0.80 |
| harbor | 0.95 | 0.98 |
| intersection | 0.65 | 0.71 |
| Medium residential | 0.64 | 0.76 |
| mobilehomepark | 0.89 | 0.91 |
| overpass | 0.81 | 0.79 |
| parkinglot | 0.93 | 0.81 |
| river | 0.72 | 0.73 |
| runway | 0.90 | 0.95 |
| Sparse residential | 0.57 | 0.69 |
| storagetanks | 0.66 | 0.60 |
| tenniscourt | 0.55 | 0.49 |

Table 5. Class-wise precision and recall report.

These obsevations indicate that the model was effective in capturing and indentifying classes that showed uniform patterns and consistent textural properties: such as agriculture, forest, chaparral, harbor etc. Model struggled with classes which had extensive intra class dissimilarity such as the baseball diamond, where each field is different: different soil texture and variability in shape and condition of the diamond. Confusion also arose when there was interclass similarities as seen in the sparse, medium and dense residential class, as our LBP descriptor does not retain the spatial information, it is easy to confuse between closely packed buildings due to scale variance.

IX. SUGGESTIONS / RECOMMENDATION

It has to be mentioned that the higher accuracy through multi-neighborhood LBPs comes at a cost of increased computation time, as we are calculating multiple LBP descriptor instead of one. Further efforts and analysis is required to optimize the time taken to compute the features. LBP can also be parallely computed to reduce the time. It would be interesting to observe trend in the trade offs between computation time and accuracy. Further to resolve difficulty in classification, we can use other feature descriptors in combination with LBP feature vector. Combining SIFT descriptors with LBP would help in tackling scale variance drawback observed in classifying dense, medium and sparse residential classes. Also, nearest neighbor classifier doesn't generalize and needs to store data, which renders it infeasible for large datasets. Hence, further experiments need to be conducted around selection of classifiers. Also, It would be interesting to observe how data augmentation affects the performance of the models.

## X. CONCLUSIONS

In this paper we proposed the use of multiple-neighborhood local binary patterns (LBPs) to classify images in the UC Merced Dataset using Nearest neighbor classifier. Dataset consisted of 21 different landuse classes ranging from baseball diamonds to building and rivers. Multi-neighborhood descriptors were formulated by concatenating singular LBPs of varying points and radius parameter combination. Performance of multi-neighborhood LBP was tested against the singular LBPs. Multi-neighborhood LBPs performed significantly better, observing more then 10% increase in overall classification accuracy. Accuracy of best performing model (77.76%) was higher than the benchmark BOVW model (76.81%) proposed in the introductory paper for the dataset. Further analysis was carried out in order to observe and understand classwise accuracy and reasons for misclassification were discussed. Our proposed model performed better in classes with consistent texture and had difficulty in differentiating classes with higher intra-class variability and inter-class similarities. Finally, further suggestion were made around improvement of the current model and adoption of other feature descriptors and classifiers to tackle the shortcoming of the current model.

## REFERENCES

[1] Pubs.usgs.gov. (2018). Cite a Website - Cite This For Me. [online] Available at: https://pubs.usgs.gov/pp/0964/report.pdf

[2] Ojala, T., Pietikainen, M., & Harwood, D. (1994, October). Performance evaluation of texture measures with classification based on Kullback discrimination of distributions. In Pattern Recognition, 1994. Vol. 1- Conference A: Computer Vision & Image Processing., Proceedings of the 12th IAPR International Conference on, Vol. 1, pp. 582-585.

[3] T. Ojala, M. Pietikåinen, and T. Måenpåå, "Multiresolution gray-scale and rotation invariant texture classification with local binary patterns," IEEE Transactions on Pattern Analysis and Machine Intelligence, vol. 24, no. 7, pp. 971–987, july 2002.

[4] Dinesh Kumar, M., Babaie, M., Zhu, S., Kalra, S., & Tizhoosh, H. R. (n.d.). A Comparative Study of CNN, BoVW and LBP for Classification of Histopathological Images.

[5] Y. Yang and S. Newsam, "Bag-of-visual-words and spatial extensions for land-use classification," in International Conference on Advances in Geographic Information Systems, 2010, pp. 270–279.

[6] S. Lazebnik, C. Schmid, and J. Ponce, "Beyond bags of features: Spatial pyramid matching for recognizing natural scene categories," in IEEE International Conference on Computer Vision and Pattern Recognition, 2006, pp. 2169–2178.

[7] A. Krizhevsky, I. Sutskever, and G. E. Hinton, "Imagenet classification with deep convolutional neural networks," in Conference on Neural Information Processing Systems, 2012, p. 10971105

[8] O.A.B. Penatti, K. Nogueira, and J.A. dos Santos, "Do deep features generalize from everyday objects to remote sensing and aerial scenes domains?," in IEEE Computer Vision and Pattern Recognition Workshops, 2015, pp. 44–51

[9] Y. Xiao, J. Wu, and J. Yuan, "mCENTRIST: A multi-channel feature generation mechanism for scene categorization," IEEE Transactions on Image Processing, vol. 23, no. 2, pp. 823–836, 2014.

[10] J. Wu and J. M. Rehg, "CENTRIST: A visual descriptor for scene categorization," IEEE Transactions on Pattern Analysis and Machine Intelligence, vol. 33, no. 8, pp. 1489–1501, august 2011

[11] A. Gunay and V. V. Nabiyev, "Automatic age classification with LBP," *2008 23rd International Symposium on Computer and Information Sciences*, Istanbul, 2008, pp. 1-4. doi: 10.1109/ISCIS.2008.4717926

[12] Castelluccio, M., Poggi, G., Sansone, C., & Verdoliva, L. (n.d.). Land Use Classification in Remote Sensing Images by Convolutional Neural Networks. Retrieved from https://arxiv.org/pdf/1508.00092.pdf